\documentclass{article}

\usepackage{times}
\usepackage{graphicx} 
\usepackage{subfigure} 
\usepackage{amsfonts}	

\usepackage{natbib}
\usepackage{amsmath}
\usepackage{algorithm}
\usepackage{algorithmic}
\usepackage{mathtools}
\usepackage{hyperref}

\usepackage[accepted]{icml2016}
\newcommand{\bs}[1]{\boldsymbol{#1}}
\newcommand{\DA}{\text{DA}}
\newcommand{\RT}{\text{RT}}
\newcommand{\stateSet}[1]{\mathcal{\mathcal{S}^{#1}}}
\newcommand{\actionSet}[1]{\mathcal{A}^{#1}}
\newcommand{\transitionSet}[1]{\mathcal{P}^{#1}}
\newcommand{\rewardSet}[1]{\mathcal{R}^{#1}}
\newcommand{\state}[2]{s_{#1}^{#2} }
\newcommand{\postState}[1]{s_{#1}^{\RT,a}}
\newcommand{\action}[2]{a_{#1}^{#2}}
\newcommand{\reward}[2]{r_{#1}^{#2}}
\newcommand{\td}{t_d}

\newcommand{\eqs}[2]{Eqs.~(\ref{#1})-(\ref{#2})}

\renewcommand{\algorithmicrequire}{\textbf{Input:}}
\renewcommand{\algorithmicensure}{\textbf{Output:}}

\newenvironment{mydescription}[1]
{\begin{list}{}%
		{\renewcommand\makelabel[1]{##1 \hfill}%
			\settowidth\labelwidth{\makelabel{#1}}%
			\setlength\leftmargin{\labelwidth}
			\addtolength\leftmargin{\labelsep}}}
	{\end{list}}

\icmltitlerunning{Hierarchical Decision Making In Electricity Grid Management}

\newcounter{egc}

\begin{document} 
	
	\twocolumn[
	\icmltitle{Hierarchical Decision Making In Electricity Grid Management}
	
	\icmlauthor{Gal Dalal}{gald@tx.technion.ac.il}
	\icmladdress{Technion, Israel}
	\icmlauthor{Elad Gilboa}{egilboa@tx.technion.ac.il}
	\icmladdress{Technion, Israel}
	\icmlauthor{Shie Mannor}{shie@ee.technion.ac.il}
	\icmladdress{Technion, Israel}
	
	\icmlkeywords{Reinforcement Learning,Power Grid, Energy}
	
	\vskip 0.3in
	]
	
\begin{abstract} 
The power grid is a complex and vital system that necessitates careful reliability management. Managing the grid is a difficult problem with multiple time scales of decision making and stochastic behavior due to renewable energy generations, variable demand and unplanned outages. Solving this problem in the face of uncertainty requires a new methodology with tractable algorithms.  In this work, we introduce a new model for hierarchical decision making in complex systems. We apply reinforcement learning (RL) methods to learn a \emph{proxy}, i.e., a level of abstraction, for real-time power grid reliability. We devise an algorithm that alternates between slow time-scale policy improvement, and fast time-scale value function approximation. We compare our results to prevailing heuristics, and show the strength of our method.

\end{abstract} 

\section{Introduction} \label{sec:introduction}
%
The power grid is a complex and vital system that requires high level of reliability. Reliability is of utmost importance, as the consequences of outages can be catastrophic. System operators (SOs) achieve reliability by means of sophisticated control operations and planning, which often require solving sequential stochastic decision problems. Sequential decision making under uncertainty in energy systems is studied in different communities such as control theory, dynamic programming, stochastic programming and robust optimization \cite{Powell2015_tutorial1,bertsimas2013adaptive,bienstock2011optimal,koutsopoulos2012optimal,bienstock2014chance}. 

Reliability assessment and control are highly complicated tasks in complex real-world systems such as the power grid. Complications in the power grid arise because of strict physical restrictions, such as generation must meet consumption continuously and transmission lines can not exceed their limited thermal capacity. Further complications stem from the structure of decision making in different time-horizons. For example, long-term system expansion and development such as building a new wind farm or a high-voltage line take years, mid-term asset management decisions such as performing maintenance are decided upon months in advance, short-term generation schedules are planned daily, and real-time operational control decisions are made on the scale of minutes. In these interdependent hierarchical decision making processes decisions are taken by multiple stakeholders. Furthermore, over the last decade, wind and solar energy sources become increasingly preeminent with further significant expansion being envisaged \cite{talbot2009lifeline}. These generators introduce high uncertainty to the system, making the control task significantly more difficult. The complex dependence between multiple time-horizon with growing uncertainty, the curse of dimensionality when dealing with large systems, and the non-linear dependence of reliability measures to the multiple time-horizon decisions, make this problem extremely hard to tackle. 

To stress the dimensionality complexity, consider the IEEE RTS-96 power network used in our experiments \cite{wong1999ieee}. This network is an example for a power grid of a medium sized European country or a state in the USA. Its state-space is $O(10^{300})$, and its action space is  $O(10^{100})$; see Sec.~\ref{sec:experiments}. Assessment of each control choice carries a computational burden as it requires solving a set of non-linear trigonometric equations named alternating current power flow (ACPF); see Sec.~\ref{sec:decisionProcessAndPowerFlow}.

Nowadays, the common practice in industry is solving large mixed integer programs (MIP), often with a linear relaxation, in an attempt to reach a valid solution \cite{grainger1994power,allan2013reliability}. Although this model is extensive, its computational burden makes it hard even for deterministic predictions (taking an order of a day in real-world systems), and inappropriate in the stochastic case.
This limits SOs to sample snapshots of future grid states or analyze a few sequential trajectories. The narrow view of possible outcomes is likely to miss important benefits and increase the costs of decisions, thereby offering little in terms of dealing with uncertainty. 

To handle uncertainty, work has been done in stochastic optimization and control theory. These often use restrictive simplifications such as independence between the decision processes in the different time-scales or consider myopic decisions only \cite{abiri2009efficient,wu2010security,abiri2013two}.  

Another approach is to use approximate dynamic programming \cite{powell2007approximate,si2004handbook}. However, the natural hierarchical structure of the problem, where several stakeholders operating in different time-scales and exposed to different information are making decisions with mutual influence, does not naturally fit the standard Markov Decision Process (MDP) structure. Furthermore, the problem is heavily constrained, since physical electrical restrictions must be met at all times.

Making this problem tractable requires a level of abstraction in the form of fast proxy methods to approximate the impact of real-time decisions on longer-term reliability and costs. 
To our knowledge, few attempts have been made to construct such proxies using tools from machine learning. An example for such, is the work conducted in a recent European project, iTesla \cite{iTesla}. This work focuses on analyzing snapshots of system states at different time points using data-mining methods. Then, classification and clustering algorithms are used for constructing security rules for predicting reliability level, given a failure and an electrical network state \cite{anil2013benchmarking}. Such approaches can aid SOs in real-time control, but lack the dynamic perspective of state-action evolution needed to evaluate consequences of policies in a sequential decision making scenario.

In this work we suggest a novel approach to mitigate the intractability of the hierarchical decision making problem of the day-ahead (DA) and real-time (RT) reliability of the power grid. The contributions of our work are:
\begin{itemize}
	\item We introduce an interleaved MDPs hierarchical structure with separate state space, action space, and reward metric.
	\vspace{-3pt}	
	\item We devise an algorithm that alternates between high-level policy improvement and lower-level value approximation, i.e., the policy improvement in the first MDP is based on the second MDP's value function. 
	\vspace{-3pt}	
	\item We show the efficacy of our method on a medium-sized power grid problem. 
	\vspace{-3pt}	
	\item We introduce a new real-world application to the RL community and provide a simulation environment.
\end{itemize}

The rest of paper is organized as follows. In Sec.~\ref{sec:background} we present background on power system engineering. In Sec.~\ref{sec:formulation}, we formulate the two-layer MDPs. In Sec.~\ref{sec:algorithm}, we introduce our interleaved approximate policy improvement (IAPI) algorithm and present results on the  IEEE RTS-96 network. We conclude our work in Sec.~\ref{sec:discussion}.

\section{Background}
\label{sec:background}
In this section we present a brief introduction to the field of power systems engineering. This is vast a field with extensive background and theory. For more information please refer to  \cite{grainger1994power,allan2013reliability}.

\subsection{Decision Processes and Power Flow in Power Grids}
\label{sec:decisionProcessAndPowerFlow}
\begin{figure}
\centering
\includegraphics[scale=0.35]{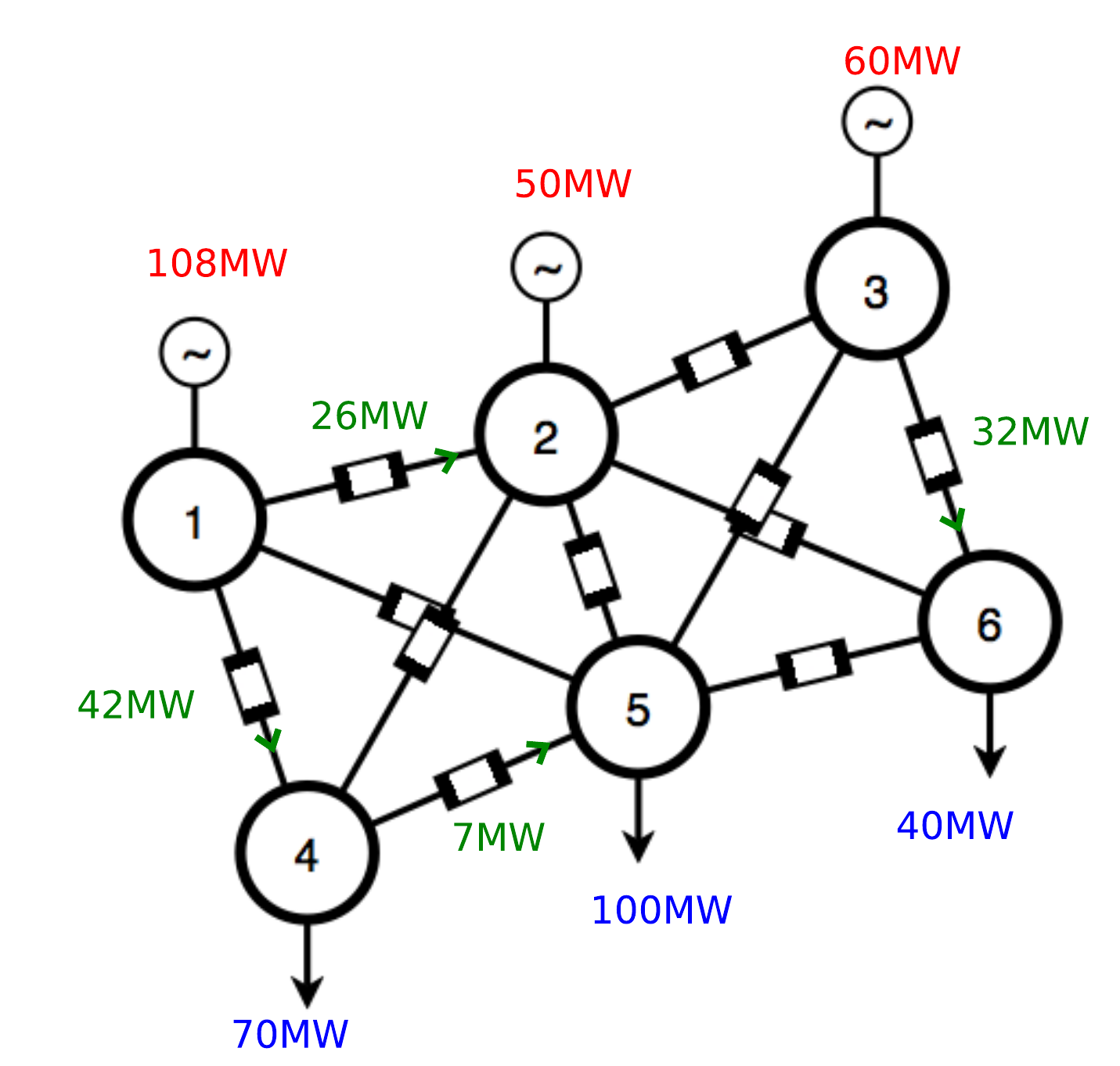}
\caption{Wood \& Wollenberg 6-bus system, with generation values in red; load values in green; and  transmission line flow values in blue, obtained from an AC power-flow solution.}
\label{case6vals}
\vspace{-0.6cm}
\end{figure}

To better explain the multiple time-horizon decision processes we use a toy 6-bus power grid example \cite{wood1996power}, shown in Fig.~\ref{case6vals}.  The 6-bus system is composed of 6 electrical nodes referred to as ``buses''. Each bus can have loads and generators attached to it. Loads (shown in blue) are consumers (e.g., large neighborhoods or cities and factories), and generators (shown in red) are power producers such as nuclear plants, coal plants, wind turbines, and solar panels. Load values change continuously throughout the day and closely follow daily, weekly, and yearly profiles. Controllable generators are operated such that the overall power generation meets the overall load at all times (up to transmission losses). The edges connecting the buses represent transmission lines which, due to thermal restrictions, can only transfer a limited amount of power before risking tripping. 

Given a snapshot of loads and generation values, and the power grid topology (buses and transmission lines), it is possible to solve the complete alternating current power flow (ACPF) equations. The ACPF is a set of non-convex trigonometric equations that model the physical electrical characteristics of the power grid, i.e., voltage magnitude and angles of each node \cite{cain2012history}. The ACPF solution includes the amount of power passing through each transmission line (shown in green in Fig.~\ref{case6vals}). 

In general, reliability of a power system is measured based on the avoidance of full or partial blackouts (both planned and unplanned) and their negative effect on social welfare. A blackout is an event where demand cannot be met. This can occur predominantly because of \emph{contingencies} (i.e., asset malfunctions) which lead to unsafe operation and may require the SOs to disconnect loads in order to avoid catastrophes. Contingencies can stem from multiple causes, such as a tree falling, lighting strike, poor maintenance or exceeding the thermal limits of a transmission line. To maintain a high reliability level at all times, the current practice of SOs is to immunize the system against a predetermined contingency list. A common choice for this list is all \emph{single} asset contingencies, resulting in the so-called $N-1$ reliability criterion.

However, contingency probabilities are difficult to obtain and their impact is hard to assess. Furthermore, the high penetration of  stochastic and often uncontrollable renewable generators, makes the planing tasks significantly harder for several reasons. First, generation must equal demand at all times. Second, multiple decision making processes are taking place simultaneously on multiple time-scales. Third, each decision process involves high dimensional decision variables, and complex non-linear \cite{Powell2015_tutorial1}, often intractable mathematical formulations.

For example, in the 6-bus system in Fig.~\ref{case6vals}, a system developer might plan to expand the system by building a new transmission line between buses 3 and 4. Expanding the grid is a long term process and a decision must be taken years in advance. However, this decision affects the future maintenance decisions, which will affect future daily planning  that in turn affects the future real-time control room operations. Ideally, the system developer should consider all possible future realizations of the environment, grid, and the decision processes in all other time horizons.


\subsection{Related work}
Several works in the literature of power systems, operational research and more recently machine learning offer approaches for solving sequential stochastic problems using dynamic programming. Of which, the majority of these works focus on energy storage \cite{lai2010approximate,xi2014stochastic,jiang2014comparison,scott2012approximate}, unit commitment  \cite{padhy2004unit,dalal2015reinforcement,ernst2007cross}, and energy market bidding strategies \cite{song2003operation,urieli2014tactex,jiang2015optimal}. To our knowledge, no work has been done to use MDPs for assessing the reliability in power grids.

For our proxy abstraction devise a hierarchical model. Hierarchical models, offer several benefits over flat models when appropriate. They can improve exploration, enable learning from fewer trials, and allow faster learning for new problems by reusing subtasks learned on previous problems \cite{dietterich1998maxq}. Standard approaches for hierarchical models include: planning with options (often referred to as skills) \cite{sutton1999between}, task hierarchy \cite{barto2003recent} and hierarchy of abstract machines \cite{parr1998reinforcement}. These models include levels of decision making that share the same state-space and a termination condition to switch between controllers. This structure does not fit our problem well where two separate decision makers run on \emph{different} state-spaces and \emph{temporal} resolutions.

\section{Problem Formulation} \label{sec:formulation}
\begin{figure}
	\centering
	\includegraphics[scale=0.55]{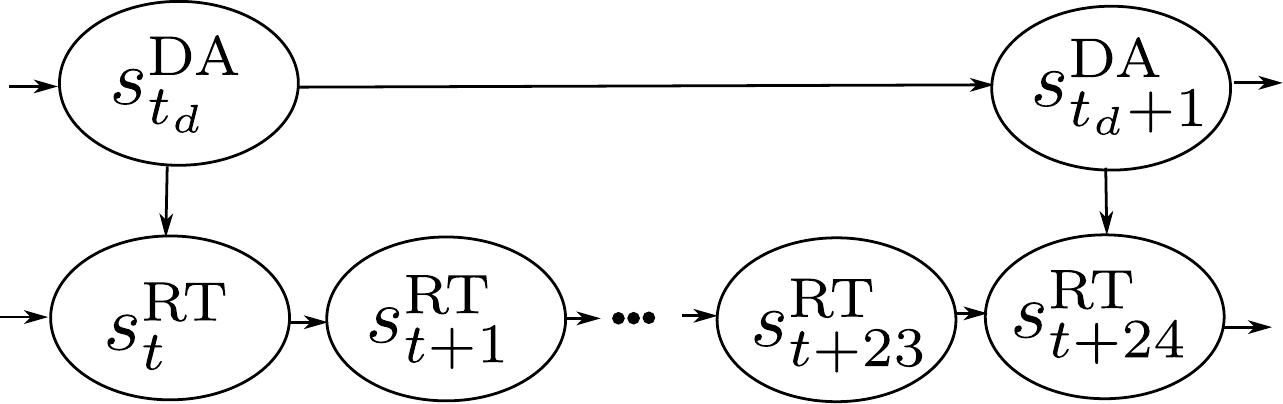}
	\caption{Day-ahead and Real-Time hierarchical MDPs. The real-time process serves as a proxy for assessing decisions taken in day-ahead process.}
	\label{fig:model}
\end{figure}

Here we present a formulation for the two sequential decision processes occurring in the day ahead (DA) and real-time (RT) in terms of a hierarchal two MDP model. DA decisions are taken in order to maximize the system's next day reliability. However, the next day reliability can only be assessed in RT, and is dependent on the system operator decision taken in RT. This results in a complex dependence between DA and RT actions and system reliability. We therefore formulate the problem using two layers of interleaved MDPs: a RT-MDP, describing the state of the system, reliability, and decisions on an hourly basis, and a DA-MDP describing the DA action of choosing a daily subset of active generators based on the upcoming day predictions. In our terminology, the former serves as a \emph{proxy} for assessing decisions taken in the latter, see Fig.~\ref{fig:model}. 


\subsection{Day-Ahead MDP}
The \DA-MDP is a tuple $(\stateSet{\DA},\actionSet{\DA},\transitionSet{\DA},\rewardSet{\DA})$. Time index is $\td$, denoting days. Day-ahead state $\state{\td}{\DA} \in \stateSet{\DA}$ consists of a day ahead prediction of hourly demand on each bus, and wind generation of each wind generator.\footnote{In this work we consider only wind generation as a renewable source for simplicity.} Therefore, $\stateSet{\DA} = \mathbb{R}^{T_{\text{D}} \cdot (n_b + n_g)}$, where $T_{\text{D}}$ is the number of intra-day time steps ($24$ in our case), and $n_b,~n_g$ are the number of buses and wind generators. For the day ahead action $\action{\td}{\DA} \in \actionSet{\DA}$ we use a simplified model which considers a binary vector indicating which generators participate in the next day's generation process. The sets of generators contained in $\actionSet{\DA}$ represent common settings an SO can choose from. This set can be constructed by experts or inferred from data. An action  $\action{\td}{\DA}$ is chosen according to a policy $\action{\td}{\DA}  = \pi^{\DA}( \state{\td}{\DA})$.  The next day state is chosen according to $\transitionSet{\DA}$, and is purely exogenous, i.e., $P(\state{\td+1}{\DA}|\state{\td}{\DA},\action{\td}{\DA}) = P(\state{\td+1}{\DA}|\state{\td}{\DA})$. The reward function $\rewardSet{\DA}$ is a complicated function of the reliability in RT. Since we cannot obtain the day ahead reward directly, we revert to use the RT reward as a surrogate for comparing DA policies. Notice that we cannot directly use the sum of RT rewards between consecutive days as a replacement for the DA reward since the model will no longer be Markovian.

\subsection{Real-Time MDP}
The RT-MDP is a tuple $(\stateSet{\RT},\actionSet{\RT},\transitionSet{\RT},\rewardSet{\RT})$. It represents the real time reliability control process. Time index is $t$, denoting intra-day time steps (e.g., hours). In RT power network operation, an operator may choose preventive actions at each time step, trying to immunize the system against potential malfunctions by attempting to avoid unreliable states. We model this decision making process using post-states \cite{powell2007approximate}, where at the beginning of each time interval, the agent observes the current state $\state{t}{\RT} \in \stateSet{\RT}$, i.e., the realized demand and wind values for this interval and chooses an action $\action{t}{\RT} \in \actionSet{\RT}$. Following the agent's action, the system is now in a post-decision state $\postState{t}$, which is the new state, after performing action $\action{t}{\RT}$ from state $\state{t}{\RT}$. Next, exogenous random information $W_t$ is obtained, informing whether equipment malfunction (contingency) occurred during time interval $t$. Given $\postState{t}$ and $W_t$, the real time reward $\reward{t}{\RT}(\postState{t},W_t)$ which represents the system's reliability, can be calculated, and a transition to $\state{t+1}{\RT}$ occurs, governed by $P(\state{t+1}{\RT}|\postState{t},W_t)$. The history of this RT process can be written as $h_t^{\RT} = (\state{0}{\RT},\action{0}{\RT},\postState{0},W_0,\reward{0}{\RT},\state{1}{\RT},\dots,\state{t}{\RT})$.
\subsubsection{Real-Time State Space}
We define a RT state $\state{t}{\RT}$ to be the tuple $(\bs{d},\bs{w},\bs{g},\bs{e})$, where:
\begin{mydescription}{$\bs{w}$}
	\vspace{-8pt}
	\item [$\bs{d}$] is a vector of stochastic nodal demand.
	\vspace{-8pt}
	\item [$\bs{w}$] is a vector of stochastic nodal wind generation. 
	\vspace{-8pt}	
	\item [$\bs{g}$] is a vector of controllable generation values. The DA action $\action{t_d}{\DA}$ determines which generators will have positive values, and which will be set to $0$ throughout the day. Each generator has minimal and maximal generation limits while in operation.
	\vspace{-8pt}	
	\item [$\bs{e}$] is the topology of the grid. Includes information of current state of each edge (transmission line). $e \in \{0,1,\dots,E\}$, where $0$ is operational and the rest is a countdown process till the line is fixed.
\end{mydescription}
\subsubsection{Real-Time Action Space}
\label{sec:realtimeaction}
A RT action $\action{t}{\RT}$ is a preventive action, that attempts to achieve better reliability of the system by immunizing against potential contingencies. The action involves redispatch $\Delta \bs{g}$, i.e., change the generation values of the working controllable generators (chosen in DA): \[\state{t}{\RT}\xrightarrow{\action{t}{\RT}} \postState{t} = (\bs{d},\bs{w},\bs{g}+\Delta \bs{g},\bs{e}).\]
Any action is allowed as long as it is within the minimal and maximal generator limits. Notice that $\bs{\Delta g}_i\ne0$ for working generators only ($\action{t_d,i}{\DA}=1$).
\subsubsection{Real-Time Transition Kernel}
The RT transition kernel can be factorized to exogenous transitions of demand, wind generation, and contingencies. It is conditioned on the last RT state and action (encoded in the RT post-state), and on the corresponding last DA decision taken to determine participating generators:
\[
\state{t+1}{\RT} = f(\postState{t},W_t,\action{t_d}{\DA}).
\] 

The dependence between RT and DA states is expressed using two sets of equations. The first is RT demand process, based on DA demand prediction: 
\begin{align}
\bs{d}_t^{\RT} = \bs{d}_t^{\DA} + \delta_t \label{eq:demand_evolution}\\
\delta_{t+1} = \delta_t + \epsilon_t, \label{eq:delta_evolution}
\end{align} 
where $\bs{d}_t^{\RT}$ is the RT demand vector at time $t$, and $\bs{d}_t^{\DA}$ is the DA prediction vector for time $t$ of the day. The dynamics in \eqs{eq:demand_evolution}{eq:delta_evolution} also hold for the wind generation process. For this work we chose this autoregressive random bias process for simplicity, however more complicated methods, such as in \cite{box2015time,papavasiliou2013multiarea,taylor2002neural},  can be considered.
The second equation coupling DA and RT determines the generators participating in current day generation process:
\begin{align}
\bs{g}_{t+1,i} = 
\begin{cases}
\bs{g}_{t,i}+\Delta \bs{g}_{t,i} \quad &\mbox{if }  i \in {\mathcal I}(\action{t_d,i}{\DA}) \\
0 \quad \quad \quad & \mbox{otherwise},
 \end{cases}
\end{align} 
where ${\mathcal I}(\action{t_d,i}{\DA})$ is the index set of generators chosen by DA action $\action{t_d,i}{\DA}$.

Lastly, random exogenous information $W_t$ specifies whether a contingency happened in the system, causing transmission line $i$ to fail, changing the network topology to $\bs{e}_{t+1}$. The probability of line $i$ to fail at each time-step is $p_i$ if at the last time-step $\bs{e}_i$ was $0$, and $0$ otherwise. 


\subsubsection{Real-Time Reward}
We choose the RT reward to be the reliability level of the power system at the current time. To assess the level of reliability, we employ the common criterion used in the industry, termed $N-1$, which assesses the system ability to withstand any contingency of a single asset.  


		
To calculate the reliability of the system, it is examined using a sequence of tests (contingency list), where each test is an attempt to take out a single line (contingency) and check if the system retains safe operation. Hence, the reward   $\reward{t}{\RT}(\postState{t},W_t)$ is  a number in $[0,1]$, expressing the portion of tests passed out of the predetermined contingency list, which includes all single contingencies $c\in{\mathcal N_{-1}}$. The reliability is calculated for a given state of the grid, and is dependent of current topology ($\bs{e}_t$) and the changes to the topology due to possible new contingencies ($W_t$) . In practice, preserving the system in safe operation means being able to obtain a feasible solution to the power flow equations (see Sec.~\ref{sec:background}) of the network circuit. We define $\mathbb{I}_{[\text{PF}(c,\postState{t},W_t)]}$ to be $1$ if a power flow solution exists, and $0$ otherwise. As a result, the RT reward is:
\begin{equation}
\reward{t}{\RT}(\postState{t},W_t) = \frac{1}{|{\mathcal N_{-1}}|}\sum_{c \in {\mathcal N_{-1}}}\mathbb{I}_{[\text{PF}(c,\postState{t},W_t)]}. \nonumber
\end{equation}

\section{Interleaved Approximate Policy Improvement} \label{sec:algorithm}
In this section we present our algorithm, called \emph{Interleaved Approximate Policy Improvement} (IAPI), presented in Alg.~\ref{alg:IAPI}, for jointly learning the RT reliability value function while searching for an optimal DA policy. We use the term \emph{interleaved} since the policy improvement in one MDP is based on the second MDP's value function. We use simulation based value learning to assess the RT reliability of the system and the cross entropy method \cite{de2005tutorial,szita2006learning} for improving the DA policy. Our method scales to large systems since it uses simple models with carefully engineered features and design to run on distributed computing. Since the algorithm is massively parallelizable, the more cores available, the faster the convergence will be.

Our goal is to find an optimal DA policy $\pi^{\DA}$, under the assumption that the RT policy $\pi^{\RT}$ is known. Henceforth, we will use $\pi$ to symbolize $\pi^{\DA}$. As explained in Sec.~\ref{sec:formulation}, reliability is not explicitly defined on the DA level and we instead use the RT value function $v^{\pi}$ as a surrogate for comparing between different DA policies. Differently than the common notation, $v^{\pi}$ denotes the RT value function, under the fixed RT policy $\pi^{\RT}$, and a DA policy $\pi$.

\begin{algorithm}
	\begin{algorithmic}[1] 
		\REQUIRE initial distribution $P_\psi^{(0)}$ for DA policy parameters
		\ENSURE optimal DA policy $\pi(\psi^*)$
		\STATE initialize ${\mathcal{S}^{RT}_{test}} = \emptyset$
		\REPEAT 
		\FOR{$i \leq N$}
		\STATE draw $\psi_i \sim P_\psi^{(k)}$
		\STATE sample $N_{\text{episodes}}$ trajectories using $\pi_i=\pi(\psi_i)$
		\STATE approximate $v^{\pi_i}$ using $\text{TD(0)}$
		\STATE add $\text{TD(0)}$ trajectories to ${\mathcal{S}^{RT}_{test}}$
		\ENDFOR
		\STATE \hspace{-6.4px} set empirical mean $\hat{v}_i^{(k)} = \sum_{s \in \mathcal{S}^{RT}_{test}} v^{\pi_i}(s)$, $\forall i \in [N]$
		\STATE \hspace{-6.4px} rank policies $\pi_i$ according to $\hat{v}_i$
		\STATE \hspace{-6.4px} use $\psi_i$  of the top percentile $\pi_i$ to update $P_\psi^{(k)}$
		\STATE \hspace{-6.4px} $k=k+1$
		\UNTIL{convergence}
	\end{algorithmic} 
	\caption{IAPI Algorithm} 
	\label{alg:IAPI} 
\end{algorithm} 

Our method includes the following components:

\paragraph{Day Ahead Policy Approximation}
We define a parametric DA policy as $\pi(\state{}{\DA};\psi) = \arg\max_{a^{\DA} \in \mathcal{A}^{\DA}} \psi^\top \Phi(\state{}{\DA},a^{\DA})$, where  $a^{\DA}$ is the day ahead action dictating which generators will be active during the day,  $\Phi(\state{}{\DA},a^{\DA})$ are features of DA state $\state{}{\DA}$ and action $a^{\DA}$. 

A plausible choice for mapping DA state $\state{}{\DA}$ to an action  $a^{\DA}$ is using multi-class classifiers. However, for large number of classes ($20$ in our experiments) these methods require a significant number of simulations for training \cite{bishop2006pattern}. Furthermore, approaches for classification-based policy learning often require obtaining multiple rollouts for all the actions from a state during the training procedure \cite{gabillon2011classification}, which in our case will result in a full value evaluation per each action and might prove overly encumbering.
To mitigate these complexities, our policy chooses the action that maximizes the inner product with $ \psi$. 

\begin{figure}
	\centering
	\includegraphics[scale=0.6]{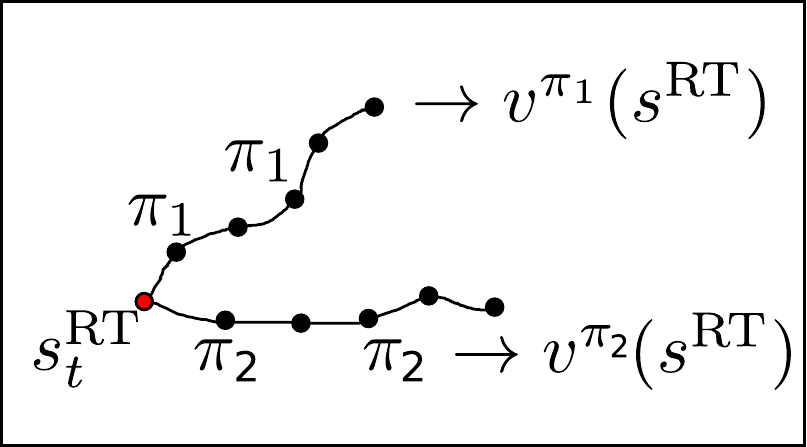}
	\caption{Day-ahead policy comparison using TD-learning of real-time value function.}
	\label{fig:valueFunctionApproximationRollouts}
\end{figure}

\paragraph{Real Time Value Function Approximation}
For a fixed DA policy $\pi$ we approximate the RT value function using the TD(0) algorithm \cite{sutton1998reinforcement}; see Fig~\ref{fig:valueFunctionApproximationRollouts}. The RT value function is parametrized as $v^{\pi}(\state{}{\RT};\theta_\pi) = \theta_{\pi}^\top \phi(\state{}{\RT})$, with the parameter vector $\theta_{\pi}$ depends on $\pi$, and $\phi(\state{}{\RT})$ being the features of RT state $\state{}{\RT}$. 

\paragraph{Day Ahead Policy Comparison}
A comparison between different DA policies $\pi_i$ is done by calculating the empirical mean of RT value function $\mathbb{E} v^{\pi_i} \approx \sum_{s \in \mathcal{S}^{\RT}_{\text{test}}} v^{\pi_i}(s)$, using a set of representative RT initial states ${\mathcal{S}^{\RT}_{\text{test}}}$. This set is composed of the full history of all RT states visited during the current IAPI iteration, enabling expected value estimation using many probable states with  only linear computational complexity in $|{\mathcal{S}^{\RT}_{\text{test}}}|$.

\paragraph{Day Ahead Policy Improvement using Cross Entropy}
DA policy improvement is achieved using the cross entropy method \cite{de2005tutorial,szita2006learning}. In this method, initial policies are sampled from a distribution $P_\psi^{(0)}$. Following which, in each iteration $k$ policy parameters $\psi$ are drawn from $P_\psi^{(k)}$, and their top percentile, according to the RT value, is used to update $P_\psi^{(k+1)}$ \cite{de2005tutorial,szita2006learning}. In our experiments we set $P_\psi^{(0)}$ such that it includes $\psi$ that equally separate $\psi^\top \Phi(\state{0}{\DA},a_{i}^{\DA})$, making this inner product equal, for all the different actions $a_{i}^{\DA}$. The distribution $P_\psi^{(k)}$ is a Gaussian mixture with means set to $\psi^{(k-1)}$ that belong to the top percentile. The convergence criterion we use in our experiments with the difference between the top-percentile values average of two consecutive iterations $ \frac{1}{N_{\text{top}}}\sum_{i=1}^{N_{\text{top}}} \left(\hat{v}_i^{(k)} - \hat{v}_i^{(k-1)}\right)^2 < \epsilon$.
By using the cross entropy method, we avoid using gradient-based optimization which may be difficult to compute in our case due to the discrete, non-linear nature of ACPF solutions and their dependence of generation \cite{cain2012history}, which dictate the level of reliability.

The criterion for comparing policies is a parametric RT value function, $v^{\pi}(\state{}{\RT};\theta_\pi)$, as oppose to using rollouts for policy evaluation \cite{gabillon2011classification}. The reason for this choice is three-fold. First, since a rollout only explores a small part of the space, assuming a structure allows us to better generalize to unvisited states. This assumption is supported by our experiments; see Fig.~\ref{fig:features_entropy}.  Second, this functional representation allows us to fairly compare different DA policies using a common set of representative RT initial states ${\mathcal{S}^{\RT}_{\text{test}}}$. Third, our end-goal is to use the value function learned by this algorithm as a proxy for system reliability in RT.

\section{Experiments}  \label{sec:experiments}
\begin{figure}
	\centering
	\includegraphics[scale=0.14]{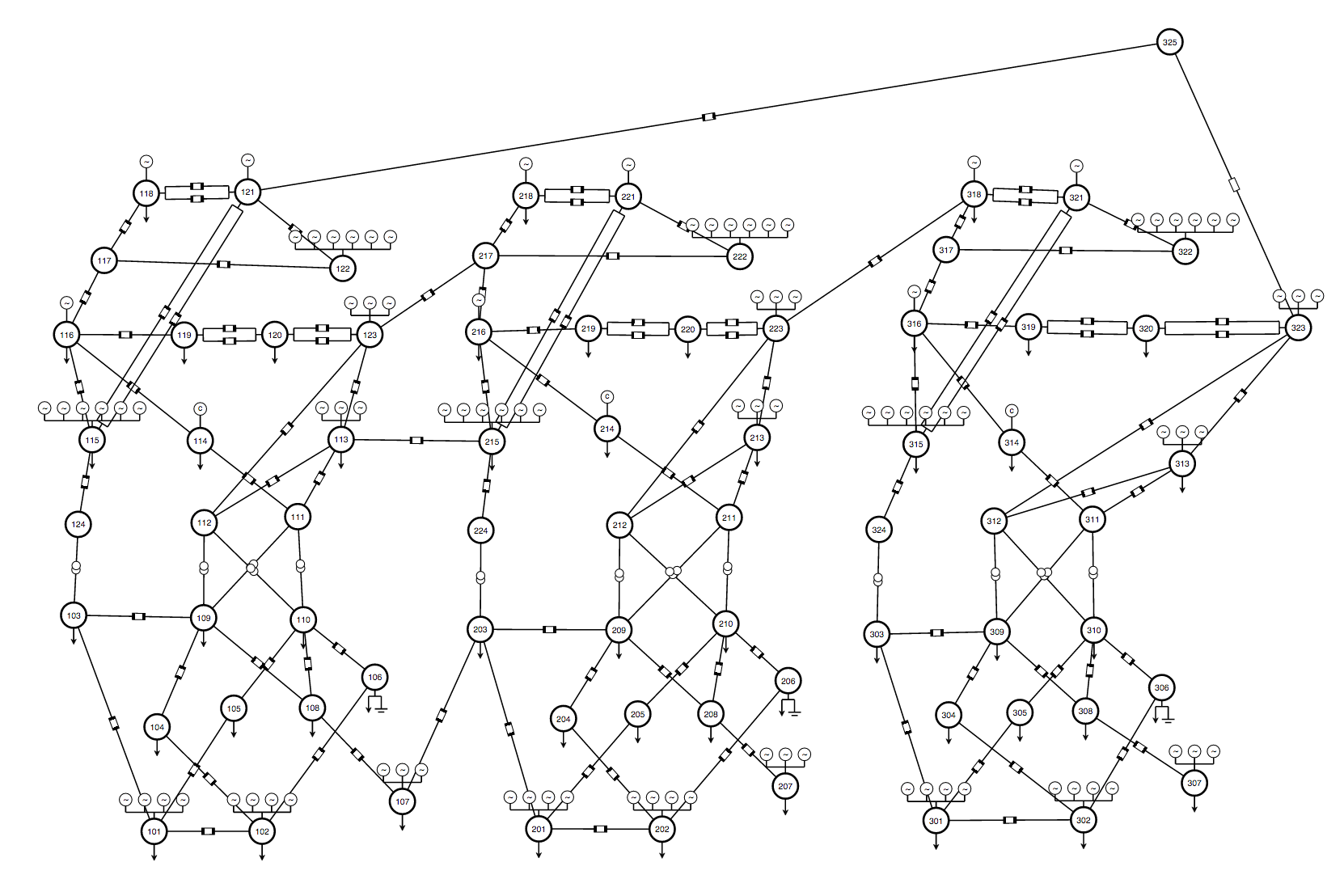}
	\caption{Diagram of the IEEE-RTS96 network we use for our experiments.}
	 \label{fig:case96}
\end{figure}

In this section we show results of IAPI algorithm on the IEEE RTS-96 test system, that is considered a standard test-case in the power systems literature \cite{wong1999ieee}; see Fig.~\ref{fig:case96}. This test-case is an example for a power grid of a medium-sized country, containing $73$ buses, $99$ generators, and $120$ transmission lines. We updated the test-case to include $9$ additional wind generators to better represent current power grids. We use daily demand and wind profiles based on real historical records as published in \cite{pandzic2015near}. As stated in Sec.~\ref{sec:introduction}, this is a complicated, high dimensional system, which cannot be solved using brute-force methods. The state space of this system can have $O(2^{120})$ line configurations, with $O(D^{73}\cdot G_\text{r}^{9})$ demand values ($D$) and wind generation values ($G_\text{r}$) at each time, which are of a stochastic nature. This is without accounting for the day-ahead prediction, which will be the power of $24$ of this number (for each hour of the day). Controlling which controllable generators are on/off makes $O(2^{99})$ integer decisions, and $O(G_\text{c}^{99\cdot 24})$ generation levels for $G_\text{c}$ possible values per each generator. 

To compose the DA action set ${\mathcal A}^{\DA}$, we define $20$ subsets of active generators chosen at random, and fix it for the rest of the simulation. These subsets contain varying numbers of generators with different capacities, to enable meeting demand for the different possible daily profiles. For the DA we use a $K+4$ feature vector
\[\Phi(\state{}{\DA},\action{}{\DA}) = (1, U_v, L_v, P, \mathbb{I}_{[\action{}{\DA} = a_1]},\dots, \mathbb{I}_{[\action{}{\DA} = a_K]}),\] where
\begin{mydescription}{$U_v$}
	\vspace{-8pt}
	\item [$K$] is the number of actions ($K=20$ in our experiments).
	\vspace{-8pt}
	\item [$U_v$] indicates if generation can meet maximal predicted daily demand. 
	\vspace{-8pt}
	\item [$L_v$] indicates if generation can meet minimal predicted daily demand.
	\vspace{-8pt}
	\item [$P$] is a barrier penalty function that penalizes if the average demand is close to the upper or lower generation bounds achieved by $a_{}^{\DA}$.
	\vspace{-8pt}
	\item [$\mathbb{I}_{[\action{}{\DA} = a_i]}$] is an indicator function over the selected DA action.
\end{mydescription}

\begin{figure}
	\centering
	\includegraphics[scale=0.3]{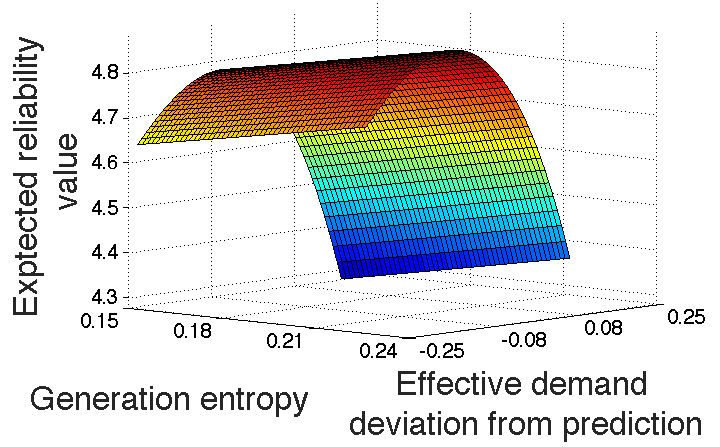} 
	\caption{Learned RT value $v^{\pi}(\state{}{\RT};\theta_\pi)$ as a function of effective demand and generation entropy across the network.} 
	\label{fig:features_entropy}
\end{figure}

For the RT policy $\pi^{\RT}$ we employ a simple heuristic, of shifting the hourly generation values to meet the realized effective demand. We consider effective demand to be demand values minus wind generation values. This is a natural approach as wind generation is not under the decision maker's control and therefore is not considered a part of regular controllable generation. The RT feature vector $\phi(\state{}{\RT})$ contains polynomial features of $(D,e_d,e_g)$, where 
\begin{mydescription}{$e_d$}
	\vspace{-8pt}
	\item [$D$] is the total RT effective demand,
	\vspace{-8pt}
	\item [$e_d$] is the demand entropy across the different buses, and
	\vspace{-8pt}
	\item [$e_g$] is the generation entropy across the different buses,
	\vspace{-8pt}
\end{mydescription}
resulting in a $10$ dimensional vector. We use the entropy feature since it compactly maps the spread of generation and demand across the network. The spread is important as the concentrations of generation and demand are directly linked to reliability issues, see Fig.~\ref{fig:features_entropy}.

For parameters for the dynamics described in \eqs{eq:demand_evolution}{eq:delta_evolution} we use \cite{lu2013comparison} and choose $\delta^w_0 \sim {\mathcal N}(0,0.05\cdot\bs{w}_0^{\RT})$ for the wind forecast error, and $\delta^d_0 \sim {\mathcal N}(0,0.01\cdot\bs{d}_0^{\RT})$ for the demand forecast error. The real time variation is chosen to be $\epsilon_t\sim {\mathcal N}(0,0.05\cdot\delta_0)$. Line failure probability $p_i$ is set to $5\cdot10^{-4}$ for each line, and its time-fill-fix $E=5$.

In our simulation we use $N_{episodes} = 50$ episodes, each with a $3$ day horizon. Each episode starts from a random DA state $\state{0}{\DA}$, drawn from several representing demand and wind profiles, to which we add normally distributed noise. The next day transition corresponds to adding a normally distributed bias to the previous day profile. In each cross-entropy iteration we evaluate $200$ DA policies ($N = 200$) and choose the top $20$-th percentile for updating $P_\psi$. The DA policies are evaluated in parallel, on a $200$ cores cluster. For the $\text{TD(0)}$ algorithm we use discounting with $\gamma = 0.95$.

In Fig.~\ref{fig:features_entropy} we show the learned RT value $v^{\pi}(\state{t}{\RT};\theta_\pi)$, as a function of the deviation of the overall effective demand (demand minus wind) from the DA prediction, and  generation entropy across the different buses. The RT value shown is marginalized over the rest of the features, time, and daily profiles. As shown in the figure, as the real-time demand deviates from the predicted demand, reliability suffers in a quadratic dependence. This is because the generators chosen in the DA will reach their upper or lower thresholds, causing generation to not meet the demand. The monotonic dependence in generation entropy implies that the higher the entropy the more reliable the system. This can be understood since high entropy corresponds to a more distributed generation throughout the network, mitigating the consequence of line outages. The reason this mitigation occurs is that when less emphasis is put on specific areas of the network, the system has more flexibility to find alternative routes from generation to demand. This, however, incurs a price in real life since generation cannot be concentrated only on cheap generators. 
%


\begin{figure}
	\centering
	\includegraphics[scale=0.22]{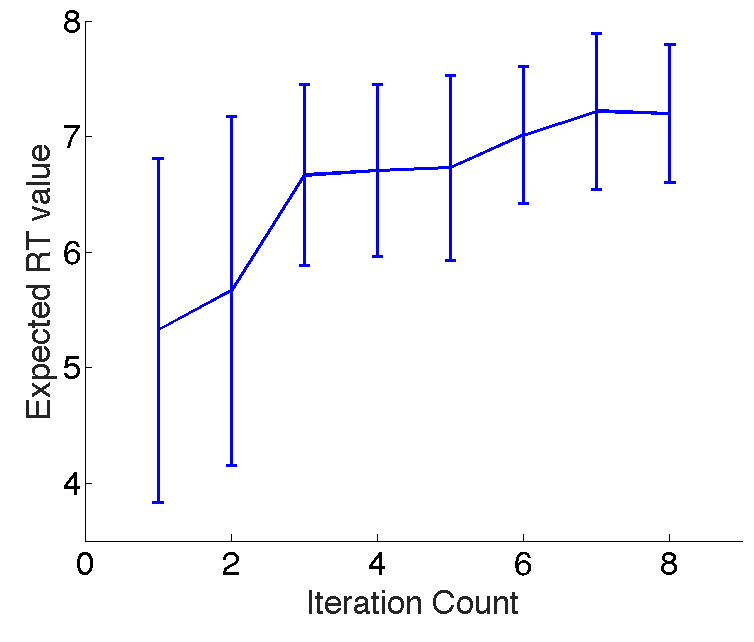}
	\caption{Convergence of the IAPI algorithm. We show the top $20$-th percentile, which is used in the algorithm to update the distribution $P_\psi$.} 
	\label{fig:convergence_value}
\end{figure}
\begin{figure}
	\centering
	\includegraphics[scale=0.4]{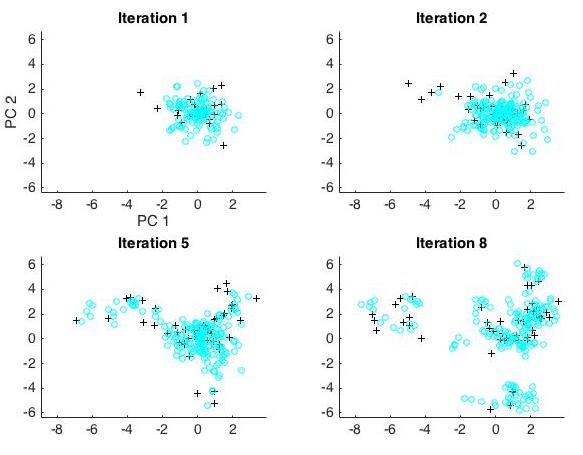}
	\caption{Projection of the top two principal components of the DA policy parameters $\psi$. The figure shows the scattering of the drawn policies parameters $\psi_i$ in each iteration, where the dark dots mark the $\psi$s corresponding to top percentile $\hat{v}_i$.} 
	\label{fig:convergence_psi}
\end{figure}
Next, in Fig. \ref{fig:convergence_value} we show the top $20$-th percentile convergence of the IAPI algorithm. As can be seen, the average value is increasing and converging after $8$ iterations, while the variance of the top percentile solutions is decreasing.  In Fig. \ref{fig:convergence_psi}, we visualize the convergence of the IAPI algorithm by projecting on the top two principal components (PC) of the DA policy parameters $\psi$. We use the same  PCs for all the plots. The figure shows the scattering of the drawn $\psi_i$ in each iteration. As described in Alg.~\ref{alg:IAPI}, each $\psi_i$ defines a policy $\pi(\psi_i)$ for which we calculate the estimated expected value $\hat{v}_i$. The dark '+' mark the $\psi$s corresponding to the top percentile of $\hat{v}_i$. As can be seen, the IAPI algorithm explores the policy space until converging to local minima.
%

\begin{figure}
	\centering
	\includegraphics[scale=0.27]{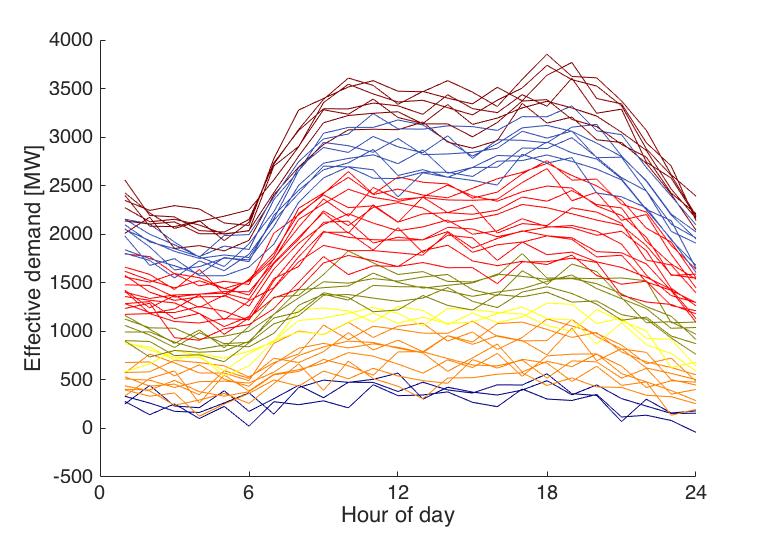}
	\caption{Daily effective demand profiles, colored according to the chosen DA action using the policy learned by the IAPI algorithm.} 
	\label{fig:action_selection}
\end{figure}

In Fig.~\ref{fig:action_selection} we present different daily effective demand profiles, colored according to the DA action chosen by the DA policy $\pi(\state{}{\DA};\psi)$, that was learned by the IAPI algorithm. A clear clustering can be observed between different daily demand profiles and the resulting action taken by the DA policy. The policy distinguishes between different consumption patterns and maps them to a corresponding set of active generators for reliable operation of the day to come.

\begin{figure}
	\centering
	\includegraphics[scale=0.2]{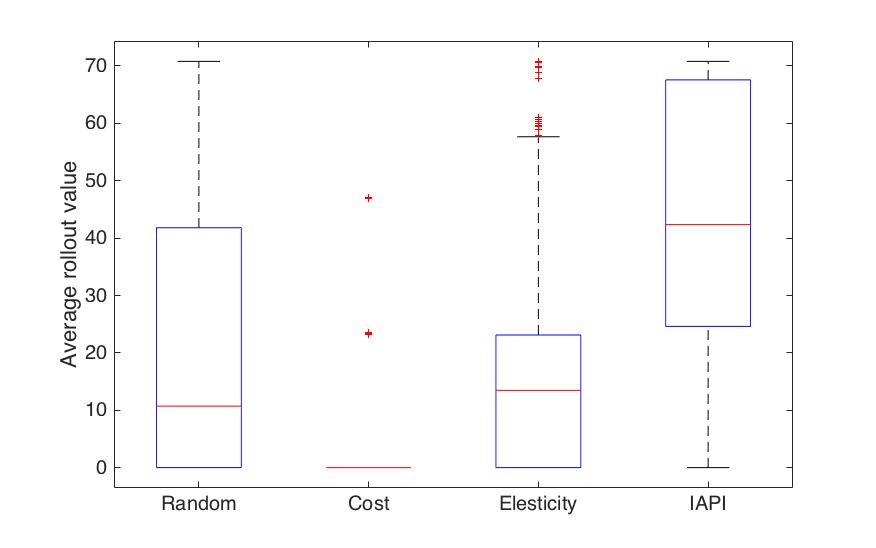} 
	\caption{Box-plot summary of the three heuristic policies and the policy learned using the IAPI algorithm. Higher is better.}  
	\label{fig:boxplot}
\end{figure}

To test our algorithm we compare the learned DA policy to three common heuristics. Taking the daily state as an input, these heuristics choose an eligible generator subset that can satisfy the maximal effective demand according to that day's DA prediction. The difference between them is the way they choose among the eligible subsets of each day. 'Random' chooses one at random, 'Cost' chooses the cheapest combination of generators, and 'Elastic' chooses the subset with the most flexible generators, having the largest ratio between upper and lower generation limits. We evaluate the performance of the different policies using rollouts of $2000$ episodes per each policy. Fig.~\ref{fig:boxplot} presents the box-plots of the results. As can be seen, the value varies greatly between the different methods. In the 'Random' policy, there is an almost flat spread, demonstrating a lack of preference for a single subset when encountering a new day. The 'Cost' and 'Elastic' policies produce a more concentrated spread, corresponding to their preference of subset choices. The policy learned using IAPI obtains higher reward than the heuristics. This result shows the IAPI algorithm's ability to learn a diverse DA policy.

%

\section{Discussion} \label{sec:discussion}
In this work we present an interleaved two-MDP model, inspired by the hierarchical decision making problem of managing power grid reliability. The IAPI algorithm presented alternates between improving the DA policy, and learning the RT reliability value. 
The IEEE RTS-96 network in our experiments is a large enough network to capture computational complexities that arise in real-world networks.

In this work we focus on the power grid, however our model can be adapted to other important applications with an hierarchical decision making structure in different time-scales where high level of reliability and sustainability is required. Examples for such applications are sewer systems, smart cities and traffic control.  
  
The coarse model presented in this work was crafted jointly as an initial step with several SOs. This work is the tip of the iceberg and many enhancements can be considered. For example, an important aspect that is not covered by it is budget consideration. Following the practice in the power system industry, reliability and money are often treated as different ``currencies''. Considering a budget will impose limitations on action selection and will complicate this problem even more. Another addition that can be made to extend the IAPI algorithm to interleave in reverse, i.e., alternating the DA improvement with improving the RT policy. Suspected drawbacks in this case are convergence problems, and the need for even more intense simulation.

Managing high reliability in stochastic complex systems, with interleaved decision making in different time horizons, is inherently difficult and results in intractable formulations. To mitigate this, there is a growing interest in the power system community to utilize proxies that will enable quick assessment of reliability for different states of the grid.
In this work we introduce new models and formulations, along with a simulation environment. Our hope is that this will provide a platform for other researchers in the community to develop and explore their own innovative methods, and will help to bring these two fields closer. The code for the simulation environment is available at \texttt{hidden to preserve anonymity}.	

\bibliography{icml16}
\bibliographystyle{icml2016}

\end{document}